\documentclass[conference]{IEEEtran}
\IEEEoverridecommandlockouts
\usepackage{subfigure} 
\usepackage{cite}
\usepackage{amsmath,amssymb,amsfonts}
\usepackage{algorithmic}
\usepackage{graphicx}
\usepackage{textcomp}
\usepackage{xcolor}
\usepackage{mathrsfs}
\usepackage{multirow}
\usepackage{makecell}
\usepackage{tabulary}
\def\BibTeX{{\rm B\kern-.05em{\sc i\kern-.025em b}\kern-.08em
    T\kern-.1667em\lower.7ex\hbox{E}\kern-.125emX}}
\begin{document}

\title{\LARGE \bf Autonomous UAV Landing System Based on Visual Navigation}

\author{
Zhixin Wu$^1$, Peng Han$^1$, Ruiwen Yao$^1$, Lei Qiao$^1$, Weidong Zhang$^1$,\\ Tielong Shen$^3$, Min Sun$^1$, Yilong Zhu$^2$, Ming Liu$^2$, Rui Fan$^{2,4}$\\
$^1$Department of Automation, Shanghai Jiao Tong University, Shanghai 200240, China.\\
\ \ $^2$Robotics Institute, the Hong Kong University of Science and Technology, Hong Kong SAR, China.\\
\ \ $^3$Devision of Mechanical Eningeering, Sophia University, Tokyo, Japan.\\
$^4$Hangzhou ATG Intelligent Technology Co. Ltd., Hangzhou, China.\\
Emails: 1642266915@qq.com, han\_ipac@sjtu.edu.cn, yaoruiwen88@foxmail.com, \\qiaolei2008114106@gmail.com, wdzhang@sjtu.edu.cn, tetu-sin@sophia.ac.jp,\\ msun@sjtu.edu.cn, \{yzhubr, eelium, eeruifan\}@ust.hk
\vspace{-1em}
}
\maketitle
\begin{abstract}
In this paper, we present an autonomous unmanned aerial vehicle (UAV) landing system based on visual navigation. We design the landmark as a topological pattern in order to enable the UAV to distinguish the landmark from the environment easily. In addition, a dynamic thresholding method is developed for image binarization to improve detection efficiency. The relative distance in the horizontal plane is calculated according to effective image information, and the relative height is obtained using a linear interpolation method. The landing experiments are performed on a static and a moving platform, respectively. The experimental results illustrate that our proposed landing system performs robustly and accurately. 
\end{abstract}

\section{Introduction}
\label{sec.introduction}
In recent years, great progress has been made in unmanned aerial vehicle (UAV) technology, and the application of UAVs has become increasingly extensive \cite{fan2019real}. Detecting and tracking moving targets, and autonomously landing on a platform are essential objectives to improve UAVs' application value. Most traditional ways of achieving these objectives rely on the combination of a global positioning system (GPS) and inertial navigation \cite{Farrell1999}. However, in many cases, such an integrated navigation system can only locate the UAV’s own position and orientation. It cannot obtain the relative position information between the UAV and the landing platform. Moreover, method based on GPS are susceptible to interference \cite{Yoo2003}, which increases the accident rate when landing in  complex and dangerous environments. In contrast to traditional methods, vision-based navigation systems have a strong anti-interference ability \cite{fan2019key}, and motion information from the external environment can be captured by visual sensors in real time. Consequently, using vision-based navigation methods to realize UAVs safe and rapid landing on moving targets has become a popular research direction.

Over the past decade, a significant amount of research on tracking and landing via visual navigation has been carried out. Lee et al. \cite{Lee2016} converted an RGB image into an HSV image and divided the regions where a marker could exist. Then, they built an object detection and tracking algorithm to allow a UAV to successfully land on a moving vehicle. Lange et al. \cite{xu2009} proposed a GPS/strapdown inertial navigation system (SINS) navigation-based, visual and inertial navigation system (INS) assisted scheme and achieved autonomous landing of a UAV. In \cite{Cocchioni2016}, an optimized Wellner algorithm, which can accurately find the contour of the marker, was applied in image processing to improve the recognition rate under uneven illumination. In \cite{Qiu2003}, binocular stereo vision \cite{fan2018road} was applied to an unmanned helicopter to achieve autonomous landing on a ship through introducing gradient information into the  Hough transform \cite{fan2016faster}. \cite{Wu2010} designed a UAV navigation system containing two cameras to guide the autonomous landing of the aircraft. In \cite{cesetti2010}, to guide a UAV to land safely, a feature-based image matching algorithm was used to identify natural landmarks selected from images. In \cite{sanchez2014}, autonomous UAV landing on ship was simulated on a motion platform with a helipad mark, and the position was calculated according to a vision-based system.

In this paper, a vision-based autonomous UAV landing system is proposed. We first design a topological pattern, consisting of squares and equilateral triangles, which can be easily identified by the UAV from the complicated surrounding environment. We then develop a dynamic thresholding method for image binarization. For example, if the recognition is successful, we will adjust the threshold slightly, and if the target has not been detected in consecutive frames, we will re-search for possible thresholds. As a result, the detection efficiency is enhanced. For the estimation of the UAV’s relative position and orientation with respect to the landing platform, we use effective information of the image to calculate the relative horizontal distance and heading angle deviation. A linear interpolation method is used to approximate the relative height. Finally, we carry out experiments on a static and a mobile platform, respectively, to evaluate the performance of the proposed UAV landing system.

\section{UAV Landing System}
\begin{figure}[!t]
	\begin{center}
		\centering
		\includegraphics[width=0.40\textwidth]{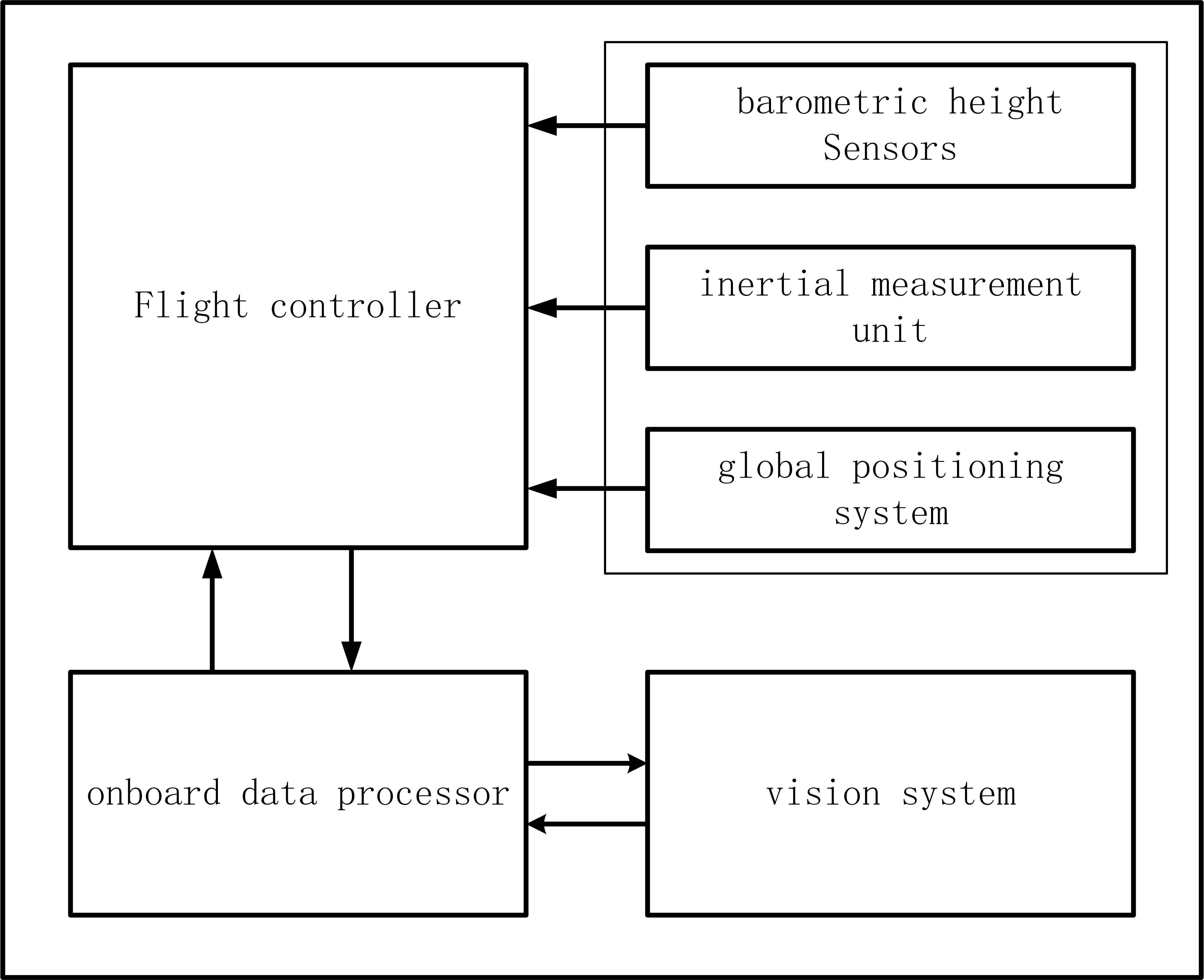}
		\caption{Block diagram of the proposed autonomous UAV landing system. }
		\label{fig.ipm}
	\end{center}
\vspace{-1em}
\end{figure}
The proposed flight system can be divided into several parts, as shown in Fig. 1. We focus mainly on the vision system, which consists of image acquisition and image processing.
In this paper, we use a Raspberry Pi 3B working in the Linux operating system as our image processing unit. Correspondingly, images are acquired by a wide-angle 175-degree HD camera named the Raspberry Pi Camera.

The availability of landmarks can affect recognition accuracy to a great extent, so we design a topological pattern, as shown in Fig. 2, which consists of squares and equilateral triangles, to act as a landmark. The outermost square of the landmark has a side length of 38 cm, and the area ratio of shapes from the outer shape to the inner shape is 32:16:4:2:1. According to this ratio, the inner four graphics are designed. The pattern has several features:
\begin{itemize}
\item  the pattern is in black and white;
\item the two outer contours are rectangles;
\item the two middle  contours are equilateral triangles;
\item the innermost contour is a rectangle.
\end{itemize}

Firstly, compared with the universal topological patterns in nature, our landmark pattern has obvious artificial characteristics, so it can be easily distinguished from the surrounding environment. Secondly, there are two kinds of shapes in the pattern and a unique area ratio between adjacent contours. The above two features can greatly reduce mis-identifications. Furthermore, the pattern can indicate UAV direction because the center of the outer square does not coincide with the triangle center. This is clearly illustrated in Fig. 2, where the red arrow starts at the center of the triangle, pointing to the center of the square. This design can provide the UAV with the yaw information with respect to the landmark. Finally, the greatest advantage of this design is the realization of hierarchical detection. When the resolution of the image is invariant, and the size or shape of the landmark in the image changes as the height of the UAV changes, the target can still be recognized. In practical applications, the quantities and position relation of squares and equilateral triangles can be modified according to the requirements. For example, the robustness of recognition at a distance can be enhanced by increasing the topological contours of the pattern.

\section{Autonomous Landing Based on Vision}
In our system, landmark detection is achieved based on the image information collected by the airborne wide-angle camera, using the OpenCV library.

\subsection{Image Preprocessing}
\label{sec.image_preprocessing}
The camera is kept vertically downward to capture images with a resolution of 640$\times$480. Acquired images are preprocessed before object detection to reduce redundant information.

To achieve this, the landmark is put on the landing platform, and the camera acquires a video stream of the landmark. For each frame, the image size is first converted to a fixed value. The image is then smoothed and filtered using a Gaussian kernel. Finally, after eliminating the Gaussian noise, the entire image is binarized into black and white.

\subsection{Dynamic Thresholding}
The principle for binarization is as follows:
\begin{equation}
{img}(x,y)=
\left\{
\begin{array}{lr}
255, & \text{if}\ \  img(x,y)>threshold \\
0, &\text{otherwise.}  
\end{array}
\right .
\label{eq1}
\end{equation}

\begin{figure}[!t]
	\begin{center}
		\centering
		\includegraphics[width=0.38\textwidth]{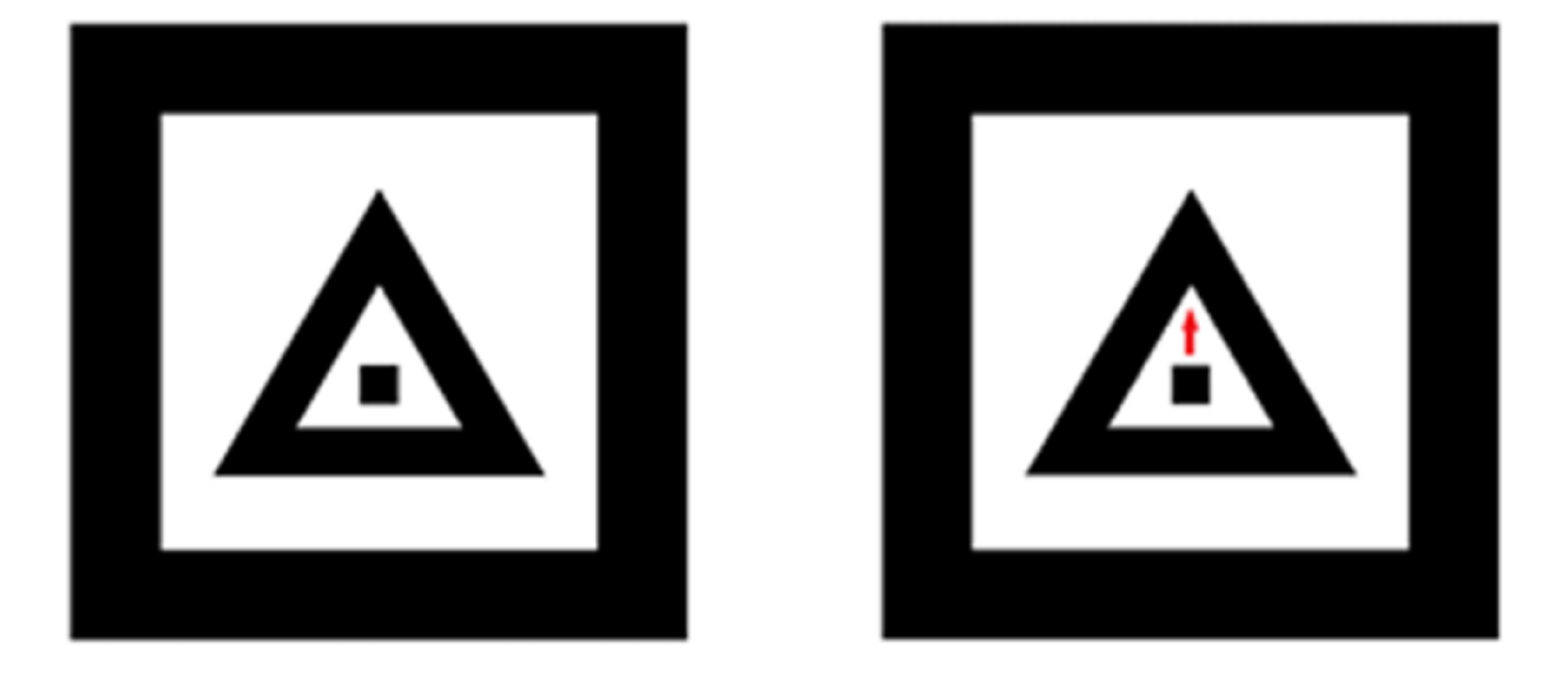}
		\caption{Designed landmarks. }
	\end{center}
\vspace{-1em}
\end{figure}
\begin{figure}[!t]
	\begin{center}
		\centering
		\includegraphics[width=0.38\textwidth]{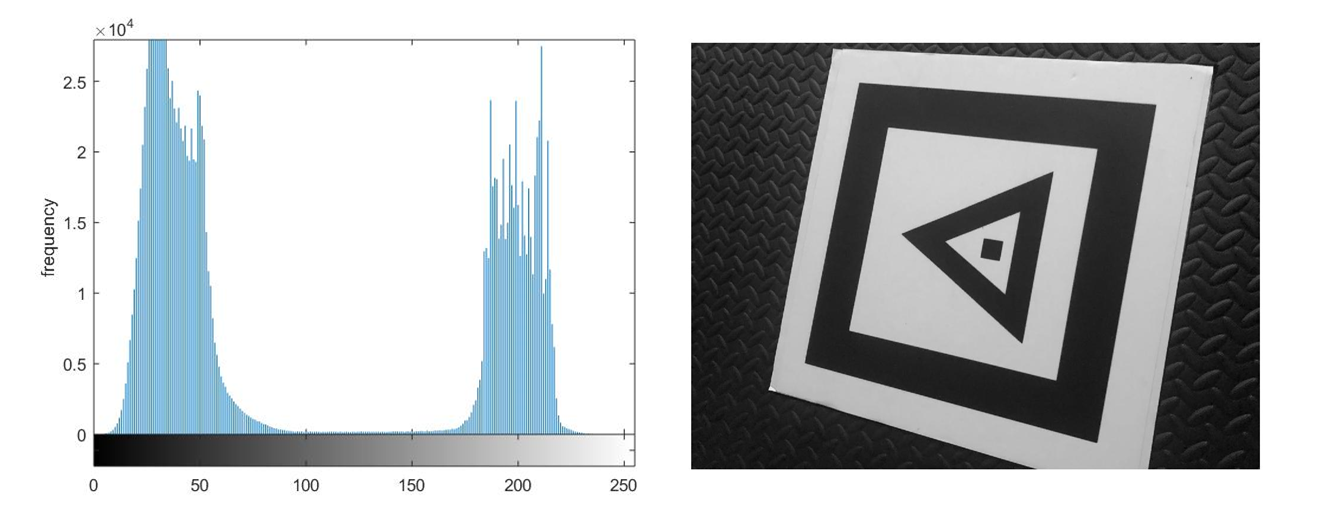}
		\caption{Gray histogram of landmark. }
	\end{center}
\vspace{-2em}
\end{figure}
In this paper, a dynamic threshold method which consists of two parts, is designed for the process of image binarization: to adjust the threshold slightly after each successful recognition and to re-search for possible thresholds if the target has not been detected in consecutive frames.

The initial value of the global threshold is set before performing object detection. Due to the motion continuity of the landmark on the image, a region of interest (ROI) is created for the target area, where the landmark is successfully detected in the previous frame. Figure 3 shows a gray histogram and a gray image of the landmark. The gray histogram is made with the concentration of gray level as abscissa and the corresponding frequency as ordinate. An obvious double peak exists in the gray histogram of the landmark; therefore, Otsu's thresholding method can be applied to find a local threshold \cite{fan2019road, fan2020pothole}. If the difference between the local threshold and the global threshold of the current frame is within a certain range, the local threshold is used for the binarization of the next frame.  

In view of possible detection failure caused by an unreasonable threshold, the corresponding action has been taken. The threshold of the image fluctuates within a large range and is initialized at the beginning of the detection algorithm. If the target is lost continuously in multiple frames, the threshold is searched in a fixed step size within this range. If the target has still not been found after being searched for for several iterations, the search range of the threshold is expanded and the search step size is reduced. In this way, the most likely range of the threshold can be traversed within a relatively short period. If the correct threshold has not been found after several iterations of traversal, the target is considered to be completely lost.

\subsection{Target Detection}
\label{sec.target_detection}
The OpenCV library function is used to retrieve all the contours of the binarized image with a hierarchical tree structure, and the correct contour is selected according to several criteria: contours’ quantities, contours’ position, contours’ area and the number of boundary inflection points. The landmark designed in this paper does not contain side-by-side sub-contours, so a hierarchical recognition algorithm can be applied to detect the target.

\begin{table}[!t]\begin{center}
				\caption{Relationship Between Quantity and Classification}

				\begin{tabular}{cc}
					
					\hline
					quantities of contours &  sequence of contours inflection point number \\
										\hline
					5 & 4433*\\
					4 & 4433, 4334\\
					3 & 443, 433, 334\\
					2 & 33, 43\\
					\hline	
				\end{tabular}
	\end{center}
\vspace{-2.5em}
\end{table}
The target marker is made up of five pure shapes nested in turn, so the total number of target contours detected must be less than five. If there is a positional relationship such as juxtaposition or coincidence, then the set of contours is excluded, and the next group will be judged. In some cases, not all contours can be found. The correct set of contours should meet the constraints on the number of contours and the number of inflection points. We encode the five contours according to the number of inflection points, which are 4, 4, 3, 3 and 4, respectively. The number of identified contours corresponding to the number of" the inflection points is shown in Table \uppercase\expandafter{\romannumeral1}.

We adopt the Douglas-Peucker algorithm to compress the closed contour curves which are divided into several segments to get the inflection points. As described in Fig. 4, the idea of this algorithm is as follows. A threshold is set in advance. First, a line can be determined between the start and endpoints on the curve. Then a point with the largest distance from the curve is chosen. If the maximum distance is smaller than the threshold, the line can be seen as an approximation of the curve. Otherwise, the curve is divided into two parts at this point. The previous process is repeated until all distances meet the threshold conditions. The polyline connected by all these segmentation points is regarded as an approximation of the contour curve. And these segmentation points are equivalent to the inflection point of the contour. 

The contents of the “sequence of contours inflection point number” column in Table \uppercase\expandafter{\romannumeral1}, such as ``4433*", are arranged according to the number of inflection points of each contour. ``*" indicates that the innermost contour is ignored during detection because the contour is too small to be recognized precisely in some cases. If the results do not match any of the conditions in Table \uppercase\expandafter{\romannumeral1}, it is judged that there is no landmark in the image area. For the cases listed in TABLE \uppercase\expandafter{\romannumeral1}, most contours are preferentially identified. When the UAV is very close to the landmark, it will start searching for the two innermost contours.

\begin{figure}[!t]
	\begin{center}
		\centering
		\includegraphics[width=0.38\textwidth]{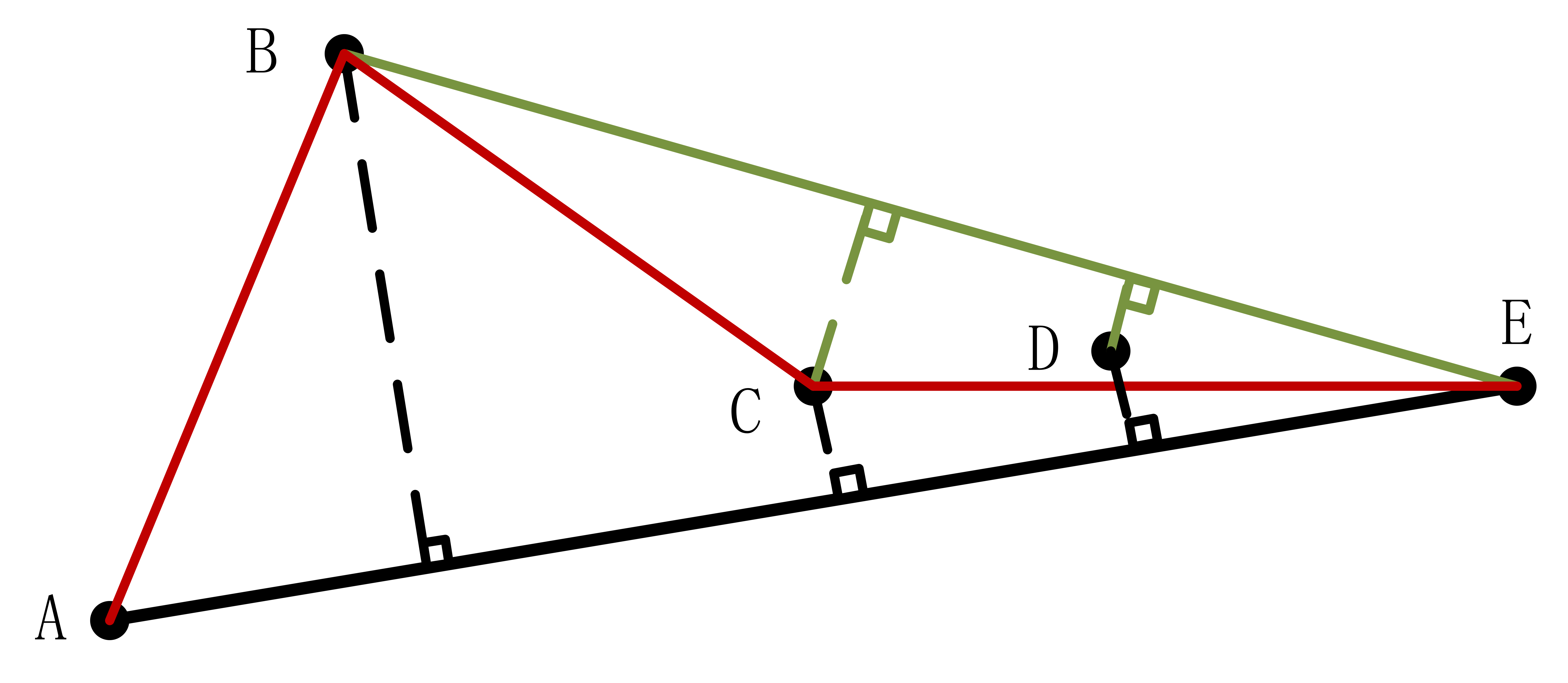}
		\caption{Douglas-Peucker algorithm. }
	\end{center}
\vspace{-2em}
\end{figure}
The area of each contour of the actual pattern is used as a priori knowledge. We can calculate the ratio of the adjacent contours area using
\begin{equation}
r_i=\frac{S_i}{S_{i+1}},
\end{equation}
where $S_i$ represents the area of the $i$-th contours of the landmark, $S_{i+1}$ represents the area of the contours inside the $(i+1)$-th contours, $r_i$ is the area ratio of the pair of adjacent contours, and $r _i$ is known in advance, so a group of contours whose area ratio deviates from the prior knowledge can be excluded. 

If a series of contours in an image satisfies all of the above constraints, they are regarded as valid information by the UAV. The entire process of target detection is shown in Fig. 5.

\subsection{Relative Position and Orientation Calculation}
After the landmark detection in the image processing, the effective feature points of the image are extracted to calculate the relative position and orientation to realize the autonomous landing of the UAV. Under the pixel coordinates system, $(x_i,y_i)$ are the pixel coordinates of the $i$-th inflection point of the larger triangle in the landmark. We can get its maximum side length $l_\text{pmax}$ and the center of the landmark $(x_l,y_l)$ using
\begin{equation}
l_\text{pmax}=\max{ \sqrt{(x_i-x_{i-1})^2+(y_i-y_{i-1})^2} }
\end{equation}
and
\begin{equation}
x_l=\frac{1}{3}\sum_{i=1}^3 x_i, \ \ \ \ y_l=\frac{1}{3}\sum_{i=1}^3 y_i.
\end{equation} 
We can then calculate the  distance under the pixel coordinates system from center of the landmark to the image center $(x_o,y_o)$ using  \cite{fan2018real}:
\begin{equation}
e_x=x_l-x_o,\ e_y=y_l-y_o,\ e_o=\sqrt{{e_x}^2+{e_y}^2},
\end{equation}
where $e_x$ and $e_y$ represent the orthogonal decomposition distance. When the side length of the equilateral triangle is $L_1$ in reality, we get the coordinates of the landmark under the image coordinates system as follows:
\begin{equation}
x_c=\frac{L_1 e_x}{l_{pmax}}, \ \ \ \ y_c=\frac{L_1 e_y}{l_{pmax}}.
\end{equation}

\begin{figure}[!t]
	\begin{center}
		\centering
		\includegraphics[width=0.43\textwidth]{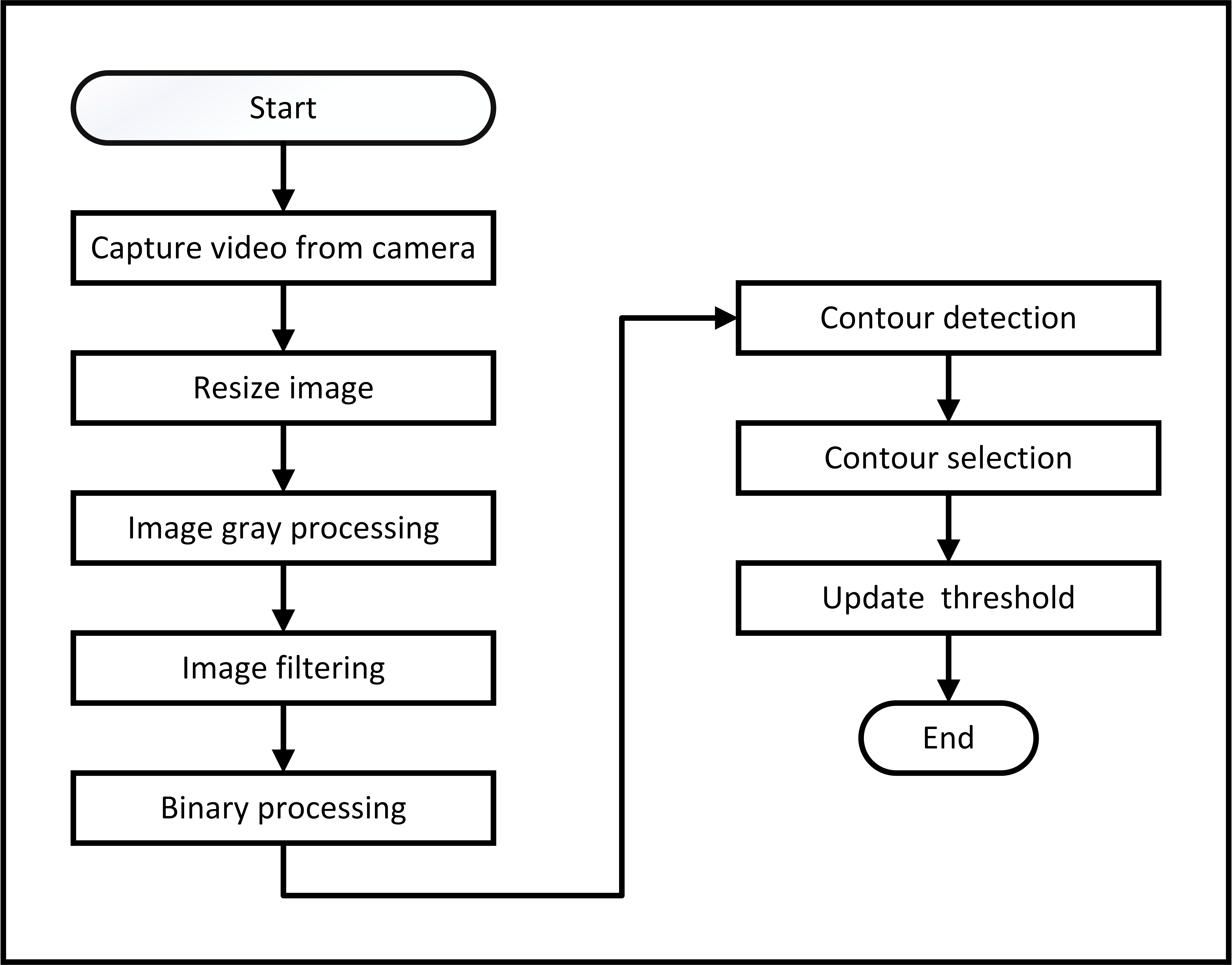}
		\caption{Target detection workflow. }
	\end{center}
\vspace{-1em}
\end{figure}

As shown in Fig. 6, $x_c$ and $y_c$ are the offsets of the landmark center from the origin of the image coordinates system, and $l_c$ is the distance from the point $(x_c,y_c)$ to the origin.

To obtain the relative position, $x_w$ and $y_w$, between the aircraft and the landing platform, the yaw angle of the UAV is defined as the angle between the flight direction and the northerly direction, denoted as $\alpha$. In the world coordinate system, we perform the coordinates transformation as follows:
\begin{equation}
\begin{bmatrix}
x_w\\y_w
\end{bmatrix}=\begin{bmatrix}
-\cos\alpha & \sin\alpha\\
-\sin\alpha & -\cos\alpha
\end{bmatrix}
\begin{bmatrix}
x_c\\y_c
\end{bmatrix}.
\end{equation}

Let $(x_{c3}, y_{c3})$ and $(x_{c4}, y_{c4})$ be the center coordinates of the triangle and the square under the image coordinates system. The vector direction from $(x_{c3}, y_{c3})$ towards $(x_{c4},y_{c4})$ indicates the expected landing direction. The angle between this vector direction and the negative direction of the $Y_c$ axis, namely the heading angle deviation between the UAV and the landing platform, $\theta_w$, can be computed using
\begin{equation}
\theta_w=\arctan\Big(\frac{x_{c4}-x_{c3}}{y_{c3}-y_{c4}}\Big).
\end{equation}

In this study, we take a linear interpolation method to approximate the relationship between the relative height $h$ and maximum side length $l_\text{pmax}$. Before the experiments, two groups of data are collected as known quantities. One group requires that the landmark’s center be aligned with the camera center, while the other group requires that the offset of the landmark’s center and the center of the camera be $D$. The measurement is performed $n$ times, and we record the UAV’s actual height  $H_i$ the $i$-th time, and also the maximum pixel side length of $L_{1i}$ and $L_{2i}$.

\begin{figure}[!t]
	\begin{center}
		\centering
		\includegraphics[width=0.30\textwidth]{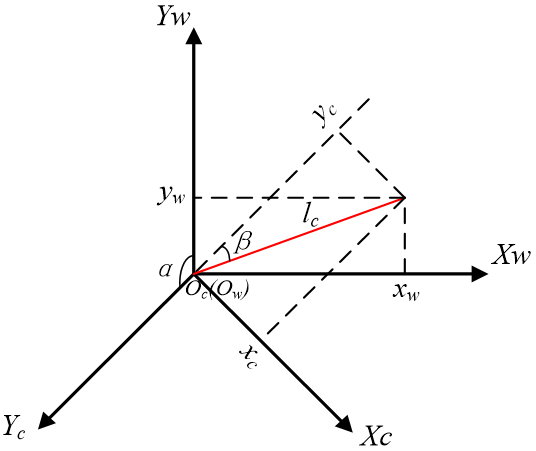}
		\caption{World and camera coordinates systems. }
	\end{center}
\vspace{-1em}
\end{figure}
The deviation between the object center and camera center will result in image distortion and pattern size change. Image undistortion is time-consuming, but it can improve the detection accuracy. In order to make the result more stable and general, we update the side length value as follows:
\begin{equation}
L_{i}=\frac{L_{2i}-L_{1i}}{D}*e_o+L_{1i},
\end{equation}
where $L_{i}$ is the corrected pixel side length, which is treated as a known quantity of linear interpolation.  The relative height $h$ of the UAV with respect to the landmark can be written as
\begin{equation}
h=\frac{H_{i+1}-H_i}{L_{i+1-L_{i}}}*\Big(
l_\text{pmax}-L_i
\Big)+H_i,L_i\textless l_\text{pmax}\leq L_{i+1}.
\end{equation}

In addition, in the process of the UAV approaching the landmark, especially when the height is less than 0.3 m, an image captured by the camera only contains part of the landmark. When only the two innermost contours are recognized, i.e., a square with the maximum pixel side length of $l_\text{pmax}'$ surrounded by a triangle with a side length of $L_2$, in reality, $l_\text{pmax}$ can be updated using
\begin{equation}
l_\text{pmax}=l_\text{pmax}'*\frac{L_1}{L_2}.
\end{equation}

                                                                                                                                                                                     After completing the position and orientation calculation, the data $x_w$, $y_w$, $h$ and $\theta_w$ are sent to the UAV control system to accomplish the tracking and landing tasks.
After completing the position and orientation calculation, the data $x_w$, $y_w$, $h$ and $\theta_w$ are sent to the UAV control system to accomplish the tracking and landing tasks.
\section{Experimental Results}
\label{sec.experimental_results}

\begin{figure}[!t]
	\begin{center}
		\centering
		\includegraphics[width=0.46\textwidth]{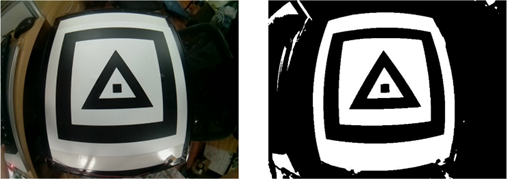}
		\caption{Original image and binary image obtained by the dynamic threshold method. }
	\end{center}
\vspace{-1em}
\end{figure}
\begin{figure}[!t]
	\begin{center}
		\centering
		\includegraphics[width=0.34\textwidth]{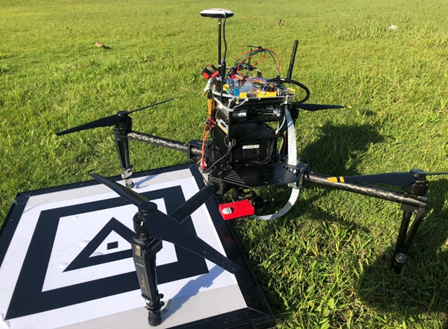}
		\caption{Static platform on lawn. }
	\end{center}
\vspace{-1.5em}
\end{figure}
In this section, we present experiments carried out on static and mobile platforms, respectively, to test the performance of the landing algorithm. We use wireless data transmission equipment to send the control commands to the UAV. Visual navigation is applied to the final stage of homing , and it  is responsible for guiding the UAV to land in a conical space with a height of 3 m from the landing platform and a diameter of about 6 m.

\begin{table}[!t]\begin{center}
		\caption{Height Corresponding To Pixel Side Length}  
		\begin{tabular}{cccccccccccc}
			
			\hline
             &1 &2 &3 &4 &5 &6 &7\\
            \hline
			$H$ &  0.3& 0.79& 1.20& 1.63& 1.99& 2.43& 2.78\\
            $L_1$ &  167.59& 63.95& 40.82& 30.02& 24.19& 19.2& 16.55\\
            $L_2$ &  159.22& 56.44& 40.31& 27.78& 21.95& 17.8& 15.65\\
			\hline	
		\end{tabular}
	\end{center}
\vspace{-2.5em}
\end{table}
The landing stage uses the Raspberry Pi 175$^\circ$ wide-angle camera to capture 640$\times$480 images at a rate of 30 fps. Image processing and target detection are all conducted on the Raspberry Pi. Fig. 7 represents an original image and binary image obtained using the dynamic threshold method presented in this paper. Before performing the experiment, we manually collect a set of fixed data about height and pixel side length for fitting the distance. These data are listed in Table \uppercase\expandafter{\romannumeral2}. Then the controller can get flight height data from the vision system in real time.

\begin{figure}[!t]
	\begin{center}
		\centering
		\includegraphics[width=0.38\textwidth]{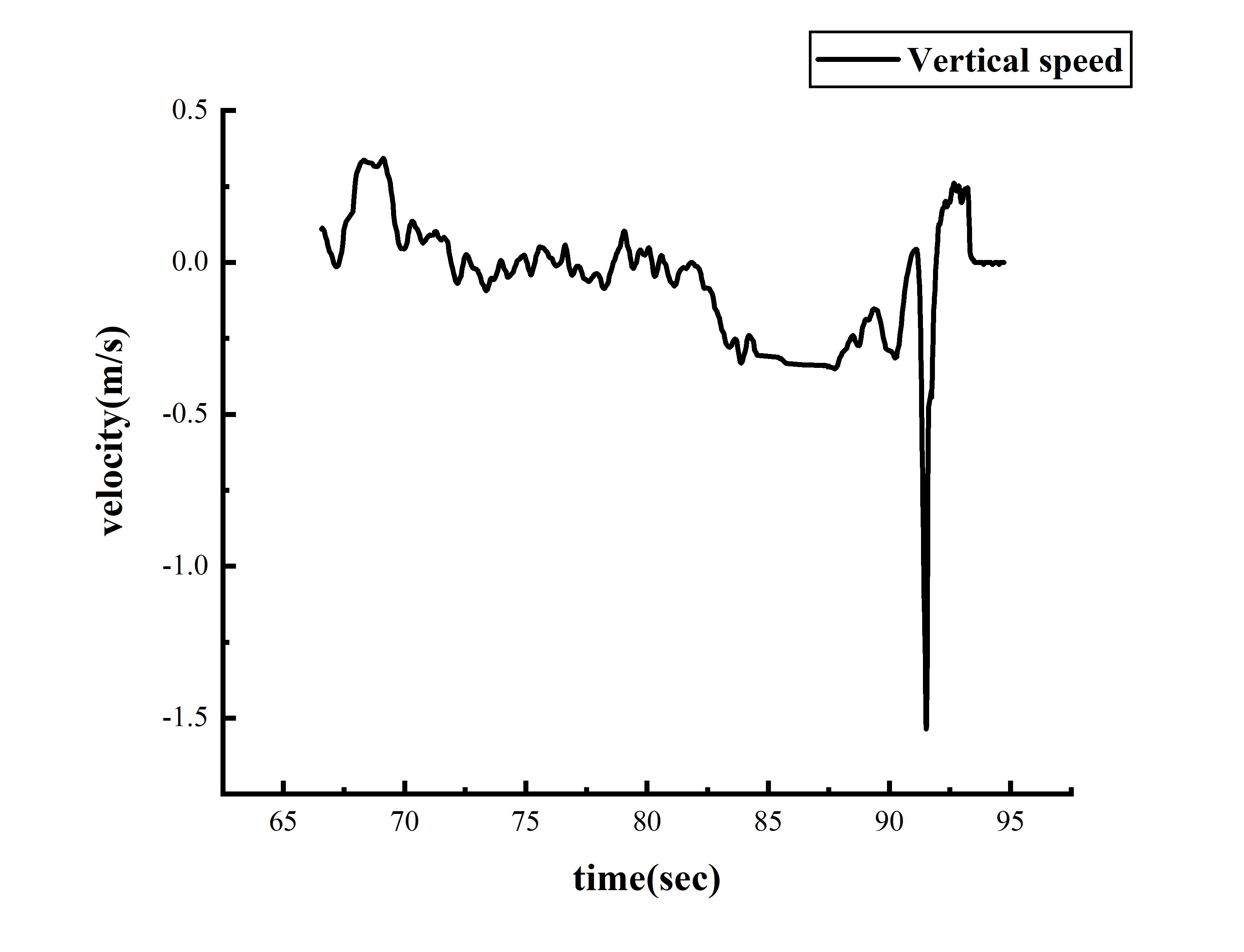}
		\caption{Vertical velocity curve of landing on static platform. }
	\end{center}
\vspace{-1em}
\end{figure}

\begin{figure}[!t]
	\begin{center}
		\centering
		\includegraphics[width=0.34\textwidth]{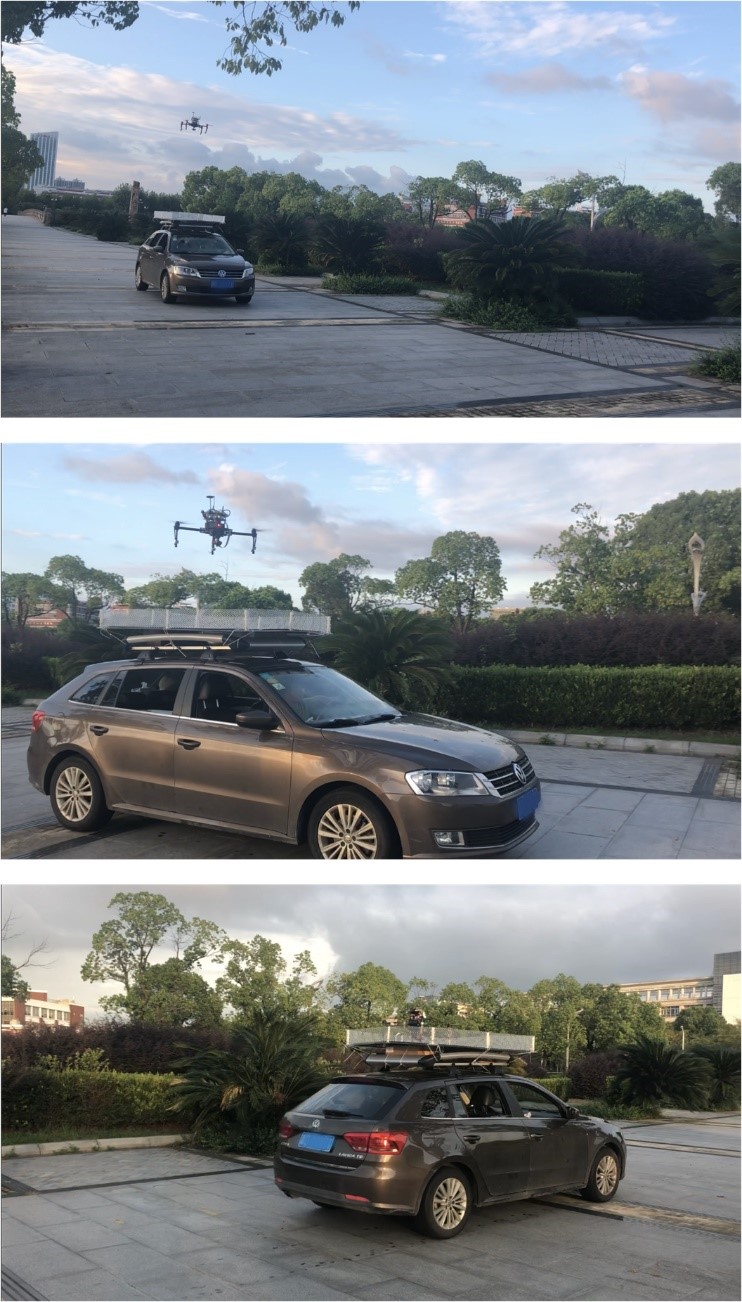}
		\caption{Landing experiment on a moving car. }
	\end{center}
\vspace{-1em}
\end{figure}
\begin{figure}[!t]
	\begin{center}
		\centering
		\includegraphics[width=0.38\textwidth]{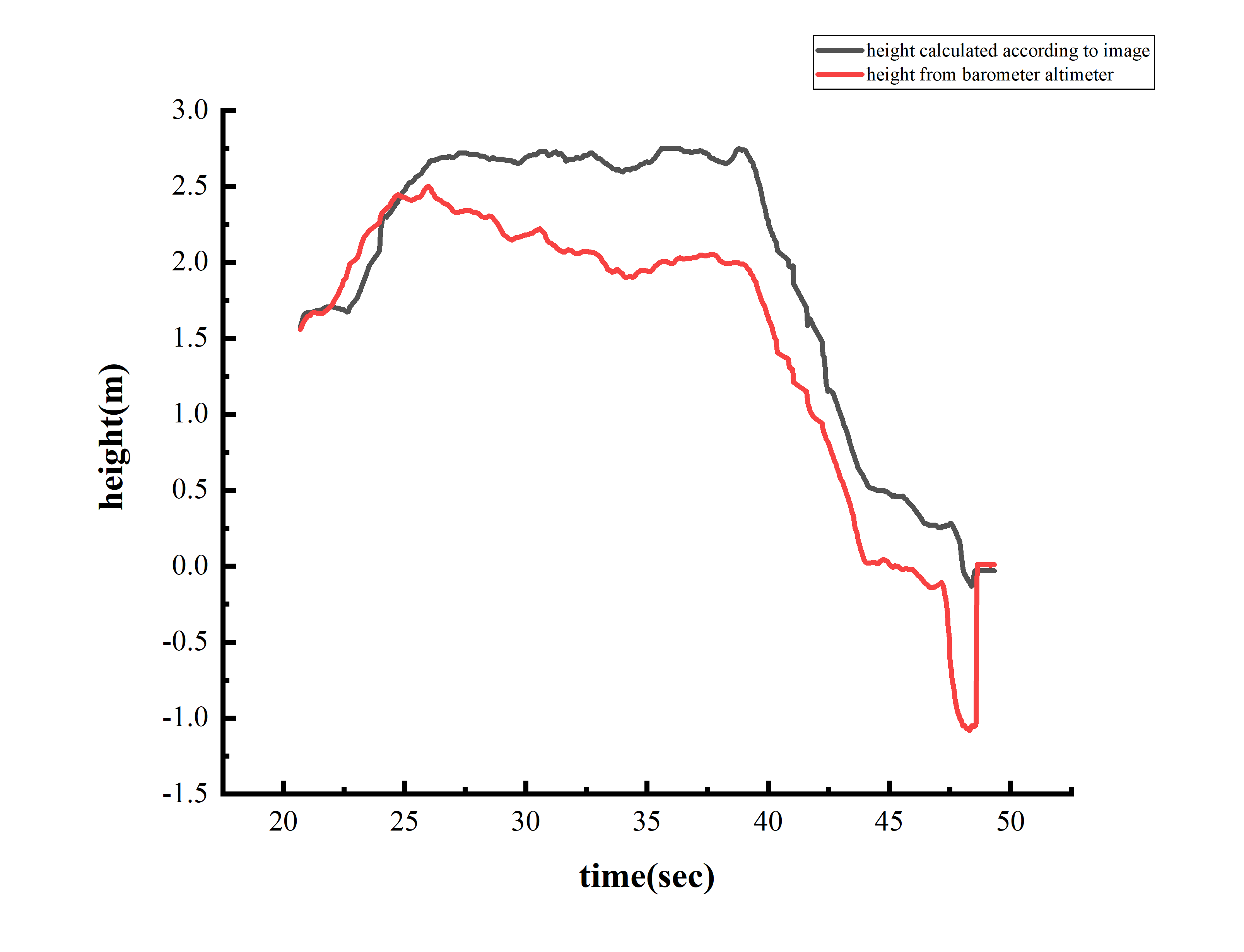}
		\caption{Vertical distance between the drone and the object. }
	\end{center}
\vspace{-1em}
\end{figure}

In the first experiment, we put the static platform on a lawn, as shown in Fig. 8. We start the experiment at 3:00 pm under bright light and slight wind. In the experiment, the camera lens is kept vertically downward, and the UAV is kept flying above the platform. Receiving commands from a mobile phone, a UAV 
performs a visual searching and hovering task then implements landing commands. The vertical velocity curve in the landing process of the UAV is shown in Fig. 9. From 66 s to 83 s, the UAV performs searching and hovering tasks, and from 83 s to 95 s, it performs the landing.

In the second experiment, we put the landing platform on a car, as shown in Fig. 10, to test the more difficult task of tracking and landing on a dynamic platform. The landmark is positioned at the center of the platform with a black background. The car is driven in a fixed direction at the speed of 13 to 14 km/h.  The UAV successfully performs the tracking and landing process with the aid of visual navigation. During the period of tracking and landing, the distance in the vertical direction between the UAV and landing platform is shown in Fig. 11. The UAV keeps tracking the target at the height of 2.5 m until it prepares to land in 36 s. The black line represents the height feedback from the UAV’s air pressure sensor, and the red line represents the height calculated by the vision system. It can be seen that the trend of change of the height calculated by the vision system is similar to that detected by the air pressure sensor, while the offset between them is mainly because the movement of the UAV limits the accuracy of the air pressure sensor, especially because the rotating blades cause strong airflow near the platform. The distance between the UAV and the landmark center in the horizontal plane is shown in Fig. 12. At the beginning of tracking, the deviation between the UAV and landmark is obvious. As time increases, the UAV moves toward the landmark, and the tracking error in the horizontal plane gradually converges to zero. The change of speed in the vertical direction is shown in Fig. 13. In the last few seconds, the speed drops suddenly to the negative direction to ensure that the UAV quickly lands on the platform.

\begin{figure}[!t]
	\begin{center}
		\centering
		\includegraphics[width=0.38\textwidth]{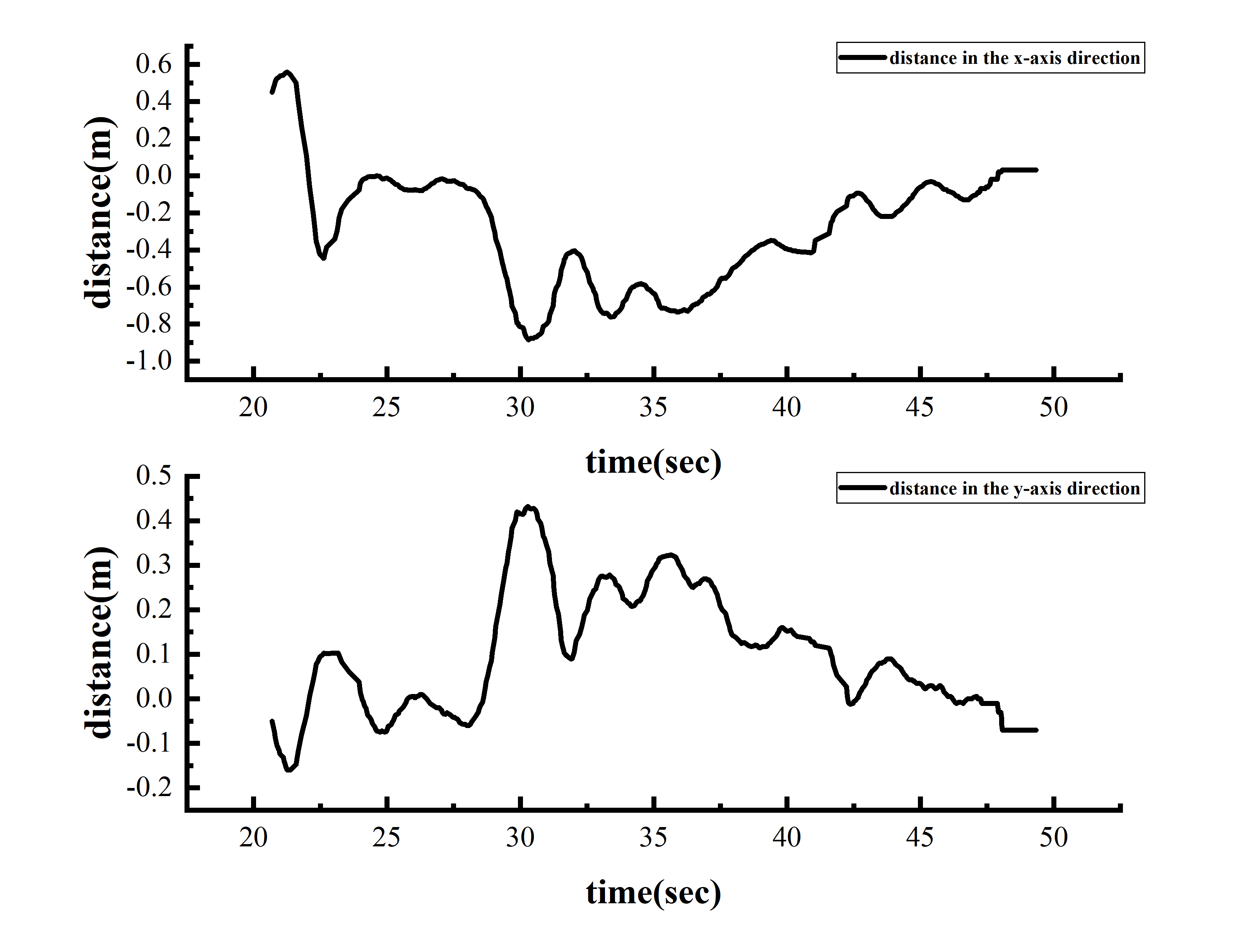}
		\caption{The distance between the drone and the moving landmark center in the horizontal plane is output from the vision detection system. }
	\end{center}
\vspace{-2em}
\end{figure}
\begin{figure}[!t]
	\begin{center}
		\centering
		\includegraphics[width=0.38\textwidth]{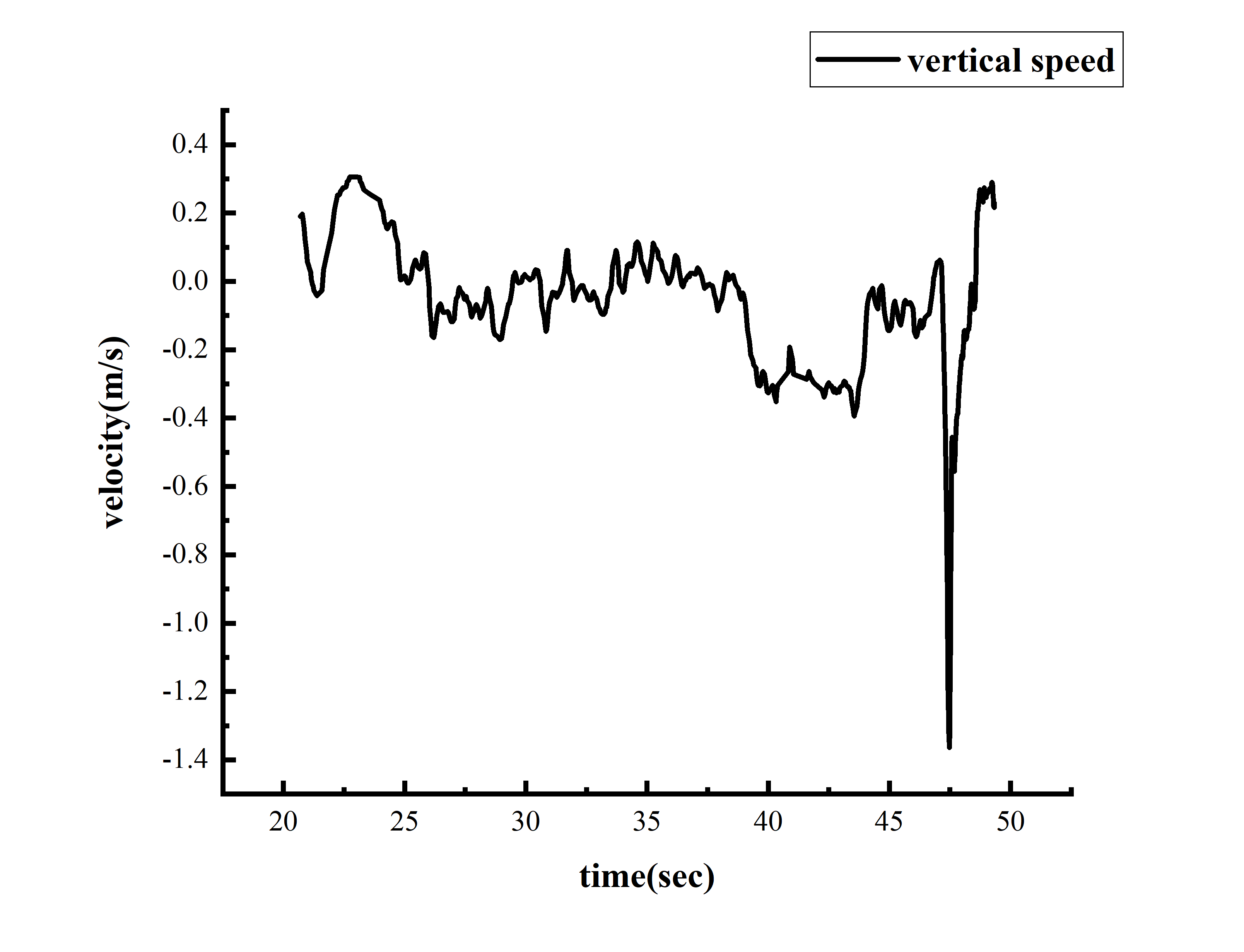}
		\caption{Vertical velocity curve for tracking and landing on moving platform. }
		\vspace{-1.5em}
	\end{center}
\end{figure}

\section{Conclusion and Future Work}
\label{sec.conclusion}
In this paper, we have designed a visual navigation method to assist UAVs in tracking moving targets and achieving autonomous landing. A well-designed topological pattern is used as a visual indicator, and the proposed approach to searching the threshold in image processing improves the efficiency of target recognition. The application of a linear interpolation method provides a precise approximation of the relative height. Experiments are successfully carried out on static and mobile platforms, showing a stable tracking and landing process. 

In the future, we will carry out more challenging collaborative experiments between unmanned surface vehicles and UAVs, including autonomous tracking and landing on the surface of water.

\section*{Acknowledgment}
This work is supported by the National Natural Science Foundation of China, under grant No. 6147318, NO. U1509211, No. 61627810, No. U1713211, National Key R\&D Program of China, under grant SQ2017YFGH001005, the Research Grants Council of the Hong Kong SAR Government, China, under Project No. 11210017, No. 21202816, and the Shenzhen Science, Technology and Innovation Commission (SZSTI) under grant JCYJ20160428154842603, awarded to Prof. Ming Liu.

\bibliographystyle{IEEEtran}

\end{document}